\documentclass{article}

\usepackage{PRIMEarxiv}

\usepackage[utf8]{inputenc} 
\usepackage[T1]{fontenc}    
\usepackage{hyperref}       
\usepackage{url}            
\usepackage{booktabs}       
\usepackage{amsfonts}       
\usepackage{nicefrac}       
\usepackage{microtype}      
\usepackage{lipsum}
\usepackage{fancyhdr}       
\usepackage{graphicx}       
\graphicspath{{media/}}     

\usepackage{amsmath}
\usepackage{booktabs} 
\usepackage{multirow} 
\usepackage{subcaption} 

\title{Hydra: An Agentic Reasoning Approach for Enhancing Adversarial Robustness and Mitigating Hallucinations in Vision-Language Models
}

\author{Chung-En (Johnny) Yu \\
    University of West Florida \\
    Pensacola, Florida, USA \\
    \texttt{cy31@students.uwf.edu} \\
    \And
    Hsuan-Chih (Neil) Chen \\
    New York University \\
    New York, New York, USA \\
    \texttt{hc4549@nyu.edu}
    \And
    Brian Jalaian \\
    University of West Florida \\
    Pensacola, Florida, USA  \\
    \texttt{bjalaian@uwf.edu} 
    \And
    Nathaniel D. Bastian \\
    United States Military Academy \\
    West Point, New York, USA  \\
    \texttt{nathaniel.bastian@westpoint.edu} 
}

\begin{document}
\maketitle

\begin{abstract}


To develop trustworthy Vision-Language Models (VLMs), it is essential to address adversarial robustness and hallucination mitigation, both of which impact factual accuracy in high-stakes applications such as defense and healthcare. Existing methods primarily focus on either adversarial defense or hallucination post-hoc correction, leaving a gap in unified robustness strategies. We introduce \textbf{Hydra}, an adaptive agentic framework that enhances plug-in VLMs through iterative reasoning, structured critiques, and cross-model verification, improving both resilience to adversarial perturbations and intrinsic model errors. Hydra employs an Action-Critique Loop, where it retrieves and critiques visual information, leveraging Chain-of-Thought (CoT) and In-Context Learning (ICL) techniques to refine outputs dynamically. Unlike static post-hoc correction methods, Hydra adapts to both adversarial manipulations and intrinsic model errors, making it robust to malicious perturbations and hallucination-related inaccuracies. We evaluate Hydra on four VLMs, three hallucination benchmarks, two adversarial attack strategies, and two adversarial defense methods, assessing performance on both clean and adversarial inputs. Results show that Hydra surpasses plug-in VLMs and state-of-the-art (SOTA) dehallucination methods, even without explicit adversarial defenses, demonstrating enhanced robustness and factual consistency. By bridging adversarial resistance and hallucination mitigation, Hydra provides a scalable, training-free solution for improving the reliability of VLMs in real-world applications.

\end{abstract}



\section{Introduction}


Vision-Language Models (VLMs) have demonstrated remarkable capabilities across various multimodal tasks, including visual question answering (VQA), image captioning, and object detection, etc. These models broadly fall into two categories: Vision-Language Pretraining (VLP) models, such as CLIP, which align images and text in a joint embedding space, and Large Vision-Language Models (LVLMs), which integrate a VLP model's visual encoder with a Large Language Model (LLM) to achieve visual conversation or complex visual reasoning tasks \cite{janowczyk2024seeing}. Currently, within the VLM research community, there is no universally agreed-upon distinction between these categories based on architectural and training paradigms. However, in the context of this paper, we adopt the commonly accepted terminology and differentiate them as follows: VLP models are pretrained on large-scale image-text pairs and can be applied to downstream visual recognition tasks without fine-tuning. This term is widely adopted in the recent adversarial attack literature. Notably, this paper focuses exclusively on image-text VLMs and does not extend to video modalities.

Despite their impressive performance, robustness remains a critical challenge in VLM research, particularly in safety-critical domains such as autonomous systems, cybersecurity, and medical AI. Robustness in this context refers to a model’s ability to resist both external adversarial attacks and internal hallucination-related errors. While prior research \cite{liu2024survey, bai2024hallucination} has primarily focused on either adversarial robustness or hallucination mitigation in isolation, a unified approach that addresses both threats simultaneously remains underexplored.


VLMs are vulnerable to adversarial attacks that manipulate either the visual input, textual input, or their image-text joint embedding space, leading to incorrect model predictions. Unlike traditional adversarial attacks in computer vision that target classification models, VLM attacks exploit the multi-modal nature of joint embeddings, subtly shifting representations to induce targeted or untargeted errors in model outputs  \cite{janowczyk2024seeing}. Recent works such as AttackVLM \cite{zhao2023evaluating} and Adversarial Illusions \cite{bagdasaryan2024adversarial} demonstrate that small, imperceptible perturbations to images can cause significant semantic shifts, misleading models into generating incorrect descriptions or answers. Notably, there is currently no advanced adversarial defense mechanism designed specifically to protect the joint embedding space of VLMs, making this a pressing challenge.


In addition to adversarial threats, LVLMs suffer from intrinsic hallucinations, where models generate incorrect, misleading, or unsupported information about an image \cite{bai2024hallucination}. Hallucinations typically arise due to biases in model parameters, co-occurrence patterns in training data, or the dominance of LLM priors over visual reasoning. For example, when presented with an image of a blue apple, an LVLM might describe it as red due to its learned linguistic biases rather than actual perception. These hallucinations can be categorized into (1) object-level, which identifies the existence or categories of objects; (2) attribute-level, which emphasizes descriptions of objects’ attributes, such as color or count; and (3) relationship-level, which assesses the relationships or relative positions among objects. Among these, object-level hallucinations—incorrectly asserting whether an object is present—are particularly critical, as they form the foundation for higher-level reasoning tasks. Thus, our work focuses on mitigating object-level hallucinations as a first step toward reducing generative errors in LVLMs.


To counteract hallucinations, prior research \cite{bai2024hallucination} has explored training-based improvements (e.g., fine-tuning on hallucination-curated datasets), decoding-based approaches, and post-hoc correction methods. Among the approches, post-hoc correction methods have shown the potential of mitigating hallucinations without intense training and adaptability of various VLM architectures. Post-hoc methods, such as Woodpecker \cite{yin2024woodpecker} and LogicCheckGPT \cite{wu2024logical}, attempt to verify and refine LVLM outputs using external tools or logical consistency checks. However, these approaches follow rigid, sequential pipelines, lacking adaptability to adversarial inputs; do not integrate adversarial robustness, meaning models corrected for hallucinations may still remain vulnerable to adversarial attacks. Given these limitations, a more adaptive, agentic approach is needed to enhance both adversarial robustness and factual accuracy.

Recent advances in LLM-based agentic reasoning, such as ReAct \cite{yao2022react} and CRITIC \cite{gou2023critic}, have demonstrated the potential of self-correcting AI agents that iteratively refine their outputs through multi-step reasoning and external tool integration. Unlike static post-hoc methods, an agentic approach enables dynamic decision-making, iterative refinement, and improved interpretability. However, to our best knowledge, no existing work has systematically explored agentic AI for both adversarial robustness and hallucination mitigation in VLMs.


To address these challenges, we propose Hydra, a training-free, agentic reasoning framework designed to enhance the robustness of plug-in LVLMs. Hydra improves factual accuracy by leveraging a structured \textbf{Action-Critique Loop}, iteratively retrieving information from multiple vision-based AI models, analyzing inconsistencies, and refining outputs to counteract both adversarial perturbations and intrinsic hallucination errors.

Our key contributions are: 
\begin{itemize}
    \item A Unified Robustness Framework: We introduce the first study that systematically integrates adversarial robustness and hallucination mitigation by evaluating LVLMs under joint embedding space attacks using hallucination benchmarks.
    \item Agentic AI for VLM Robustness: We propose Hydra, an adaptive single-agent framework that employs structured reasoning, in-context learning (ICL), and chain-of-thought (CoT) techniques to refine LVLM outputs, enhancing resilience against both adversarial attacks and hallucinations.
    \item Empirical Validation Across Multiple VLMs: We conduct extensive experiments across four LVLMs, three hallucination benchmarks, two adversarial attack techniques, and two adversarial defense strategies, demonstrating that Hydra outperforms SOTA dehallucination methods while remaining robust against adversarial perturbations.
\end{itemize}

\begin{figure}
    \centering
    \includegraphics[width=0.5\linewidth]{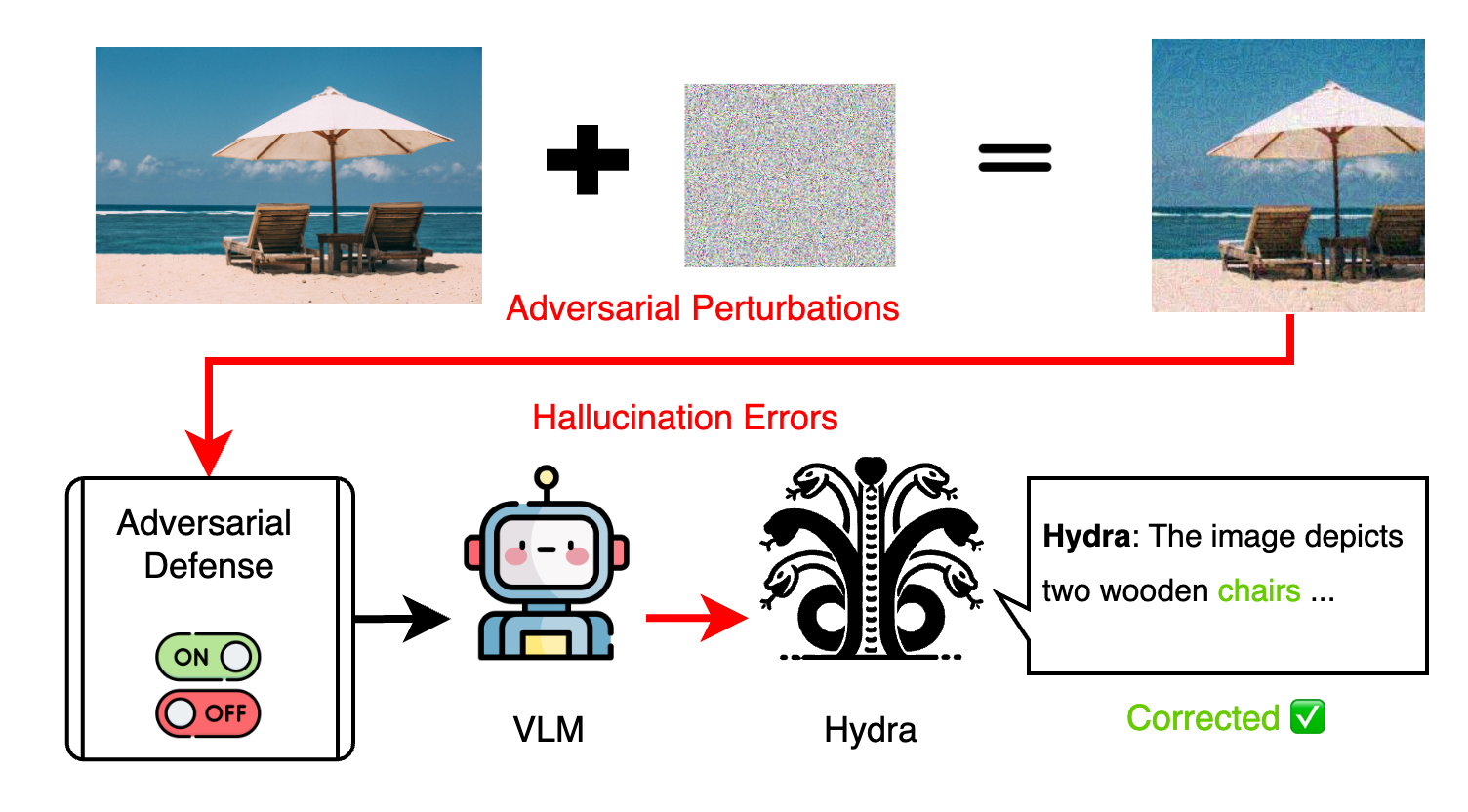}
    \caption{An overview of Hydra improving VLM robustness by addressing hallucination errors and adversarial perturbations.}
    \label{fig:hydra_overview}
\end{figure}

By integrating reasoning-driven correction with adversarial robustness, Hydra bridges the gap between VLM security and factual reliability, paving the way for more interpretable, trustworthy, and resilient vision-language models, illustrated in Figure \ref{fig:hydra_overview}.

\section{Related Works}

\subsection{Adversarial Attacks in VLMs' Joint Embedding Space}


VLMs align visual and textual representations in a shared embedding space, making them vulnerable to adversarial attacks that disrupt this alignment. Unlike conventional adversarial threats that target classification models, these attacks manipulate multi-modal embeddings, leading to semantic misalignment and incorrect predictions \cite{janowczyk2024seeing}. Recent works such as Adversarial Illusions \cite{bagdasaryan2024adversarial} and AttackVLM \cite{zhao2023evaluating} demonstrate that minor perturbations to images can significantly alter textual outputs by distorting cosine similarity in the embedding space. Notably, these attacks transfer effectively across models with similar visual encoders, such like CLIP family, posing a broader security risk.

Existing defenses primarily rely on adversarial training or preprocessing techniques like JPEG compression \cite{dziugaite2016study} and feature squeezing \cite{xu2017feature}, which, while mitigating certain attacks, often degrade performance on clean inputs and remain ineffective against hallucinations \cite{chakraborty2018adversarial}. While Adversarial Prompt Tuning (APT) \cite{li2024one} has been proposed as a defense targeting the joint embedding space, it focuses specifically on countering adversarial perturbations by optimizing textual embeddings, leaving visual representations vulnerable to attack. Moreover, APT does not address broader challenges such as mitigating hallucinations in VLMs, which are equally critical for ensuring reliable multimodal alignment.

\subsection{Hallucination Mitigation in VLMs}


Beyond adversarial threats, LVLMs frequently generate hallucinations, where models output incorrect or unsupported descriptions due to biases in training data, linguistic priors, or co-occurrence correlations. These errors are particularly concerning in object-level hallucinations, where models falsely assert the presence of objects not in the image. Addressing this issue is critical, as object-level accuracy underpins higher-level visual reasoning tasks \cite{liu2024survey}.

To mitigate hallucinations, post-hoc correction methods have been proposed. Woodpecker \cite{yin2024woodpecker} refines model outputs by extracting key objects, verifying them using external tools such as object detectors and VQA models, and revising inconsistencies. However, this approach requires multiple inference passes, making it computationally expensive. LogicCheckGPT \cite{wu2024logical} adopts a different strategy, using self-consistency checks where an LVLM is prompted with logically related questions to detect contradictions. While more efficient, this method lacks external validation, making it susceptible to intrinsic biases. Both approaches, though effective in hallucination mitigation, fail to account for adversarial robustness, as they follow static, linear correction pipelines that cannot adapt to adversarially manipulated inputs.

\subsection{Agentic AI for Reasoning and Robustness Enhancement}


Recent advances in agentic AI suggest an alternative paradigm where models refine their outputs through iterative reasoning and external tool interactions \cite{durante2024interactive, pan2023automatically}. Frameworks such as ReAct \cite{yao2022react} and CRITIC \cite{gou2023critic} integrate structured decision-making with external verification, demonstrating improvements in factual accuracy by dynamically adjusting outputs based on retrieved information. However, these methods have yet to be applied systematically to VLM robustness, particularly in the context of both adversarial defense and hallucination reduction. The ability of agentic systems to autonomously critique and refine their decisions presents a promising avenue for enhancing the reliability of VLMs.

\subsection{Research Gaps and Motivation for Hydra}


Despite progress in adversarial robustness, hallucination mitigation, and agentic reasoning, no existing agentic approach systematically addresses both adversarial and intrinsic errors in LVLMs. Current defenses either focus on external adversarial threats or post-hoc hallucination correction, but fail to integrate both by agentic framework. Furthermore, there is no advanced defense mechanism designed to protect the joint embedding space in multimodal models. Existing post-hoc hallucination correction methods lack adaptability, following predefined correction pipelines that are ineffective against adversarially altered inputs. While agentic AI has shown promise in structured reasoning tasks, it remains unexplored as a solution for both adversarial and generative robustness in VLMs.

To bridge this gap, we introduce Hydra, an agentic reasoning framework that enhances the robustness of plug-in LVLMs by integrating structured verification, adaptive critique loops, and multimodal evidence aggregation. Unlike prior methods, Hydra dynamically refines outputs by leveraging multiple visual AI models (both DL-based and transformer-based models) alongside reasoning-driven corrections, enabling it to resist both adversarial perturbations and hallucination-related errors. Our experiments demonstrate that Hydra not only outperforms state-of-the-art dehallucination methods but also improves robustness under adversarial conditions, establishing a unified approach to securing VLMs against both external manipulations and intrinsic inaccuracies.

\section{Method}

\begin{figure}
    \centering
    \includegraphics[scale=0.13]{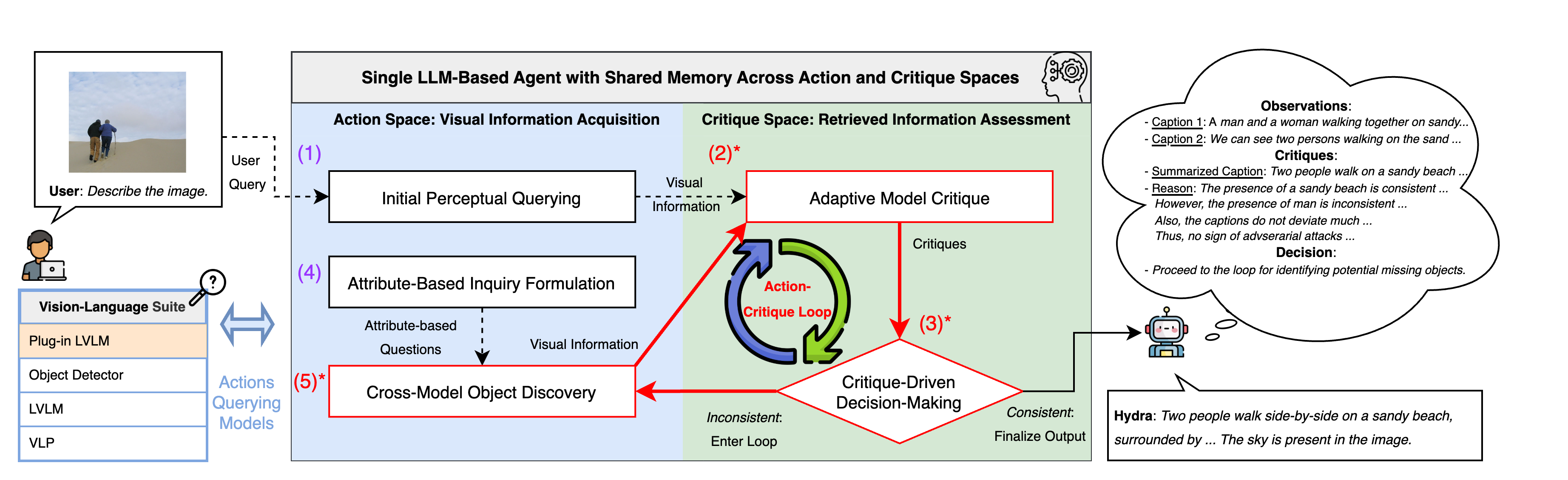}
    \caption{The Hydra framework enhances VLM robustness through an iterative Action-Critique Loop, integrating perceptual querying, cross-model object discovery, and adaptive critique to refine outputs and mitigate hallucinations.}
    \label{fig:hydra}
\end{figure}


\subsection{Hydra Architecture Overview}

Hydra is an adaptive agentic framework designed to enhance the robustness of plug-in LVLMs by iteratively refining their outputs through structured reasoning and cross-model information retrieval. As illustrated in Fig. \ref{fig:hydra}, at the core of Hydra, a single LLM-based agent interacts with a vision-language suite consisting of multiple visual AI models, such as object detectors and VLMs. With the diverse visual models within the suite, it lowers the risk of compromising to the adversarial attacks compared to a single model. This agent operates within two distinct behavioral spaces: (1) \textbf{Action Space}, where it queries visual AI models to retrieve relevant information, and (2) \textbf{Critique Space}, where it assesses and refines the retrieved information and autonomously determines the next steps. The agent maintains a shared memory space, where all retrieved information and critiques are stored, enabling informed decision-making over iterative refinements. Hydra’s workflow is structured into five subtasks:
(1) \textbf{Initial Perceptual Querying}, where the agent queries the plug-in LVLM and auxiliary vision models, such as object detectors, to extract foundational visual information; 
(2) \textbf{Adaptive Model Critique}, where the agent evaluates the retrieved visual information, generating structured critiques that assess consistency, factual accuracy, and potential hallucinations;
(3) \textbf{Critique-Driven Decision-Making}, based on the critiques, the agent determines whether further information retrieval is necessary or if a final decision can be made;
(4) \textbf{Attribute-Based Inquiry Formulation}, if additional verification is required, the agent formulates attribute-based questions derived from retrieved descriptions, enhancing object discovery.
(5) \textbf{Cross-Model Object Discovery}, using attribute-based queries, the agent iteratively retrieves responses from multiple models, aggregating evidence to improve factual accuracy. A key mechanism enabling Hydra’s adaptivity is \textbf{Action-Critique Loop}, an iterative process where the agent refines responses dynamically. If retrieved information is inconsistent, the agent formulates additional queries, critiques new responses, and re-evaluates its decision. The loop continues until a confident conclusion is reached or a predefined iteration limit is met, ensuring resilience against adversarial perturbations and hallucination errors.

Inspired by prior works, Hydra is designed as a modular, plug-and-play framework that operates independently of specific VLM architectures, leveraging fundamental reasoning techniques such as CoT and ICL to enhance robustness. Its iterative refinement process allows it to adapt dynamically to different adversarial conditions and hallucination scenarios, making it more flexible than static post-hoc correction methods. Hydra’s design mirrors its namesake—multiple adaptive components working in coordination to ensure robustness. By integrating diverse vision-language models and iterative reasoning, the system remains resilient to perturbations, compensating for failures through redundant verification. 



\paragraph{Hydra in VQA Task.}

For VQA tasks, object-level hallucination benchmarks typically have query like ``\textit{Is there \{TARGET OBJECT\} in the image?}", expecting a binary (``\textit{Yes/No}") response. Hydra mitigates object-level hallucinations by verifying object presence using multiple vision models. The workflow follows these structured steps:
\begin{enumerate}
    \item Initial Querying: The agent queries the plug-in LVLM for a detailed caption and an object detector for a list of detected objects. The target object from the user query is extracted and stored in memory.
    \item Adaptive Model Critique: The agent evaluates object presence based on both the object detector’s and the LVLM’s response, generating critiques that include a binary existence decision and supporting rationale.
    \item Critique-Driven Decision-Making: If the two critiques are consistent, the agent finalizes the answer. If inconsistent, the agent enters the Action-Critique Loop to seek further verification.
    \item Attribute-Based Inquiry: The agent formulates additional verification questions based on object attributes (e.g., ``\textit{What objects are red in the image?}"), extracting key descriptors from previous responses.
    \item Cross-Model Object Discovery: The agent queries the plug-in LVLM and two other VLMs using attribute-based questions, iteratively refining the decision.
\end{enumerate}

\paragraph{Hydra in Captioning Task.}

Hydra aims to refine captions that accurately describe all objects present in the image and remain robust against adversarial attacks. The workflow follows these structured steps:
\begin{enumerate}
    \item Initial Perceptual Querying: The agent queries the plug-in LVLM along with additional LVLMs to generate multiple image captions.
    \item Adaptive Model Critique: The agent critiques retrieved captions to identify consistent elements and potential compromised models, then summarizes a caption.
    \item Critique-Driven Decision-Making: If retrieved captions contain conflicting descriptions, the agent enters the Action-Critique Loop to refine the summary caption.
    \item Attribute-Based Inquiry: Unlike VQA, which has explicit object in the query, captioning requires the agent to extract objects in the retrieved captions. Objects missing from the summarized caption but present in multiple retrieved captions are flagged for verification.
    \item Cross-Model Object Discovery: The agent iteratively verifies flagged objects by querying multiple LVLMs and a VLP model. If validated, the object is added to the summary caption.
\end{enumerate}

Hydra’s design leverages multiple visual AI models, each with distinct visual encoders, reducing susceptibility to adversarial attacks. Additionally, its structured reasoning mitigates hallucinations by cross-verifying object presence. The training-free agentic approach also enhances interpretability, allowing human users to understand model behaviors and decision rationales.

\section{Experiment}

\subsection{Experiment Setup}

\subsubsection{Robustness Evaluation Metrics}

To evaluate the robustness of Hydra against both adversarial perturbations and intrinsic model errors, we employ object-level hallucination benchmarks. As a baseline, we assess Hydra’s performance on standard hallucination benchmarks and then introduce two SOTA adversarial attacks designed to exploit vulnerabilities in LVLMs’ joint embedding space, targeting the image modality within hallucination benchmarks, following prior studies \cite{kim2024doubly, yin2023vlattack}.

On the defense side, we apply two widely used adversarial countermeasures \cite{dziugaite2016study, xu2017feature} based on image preprocessing techniques. Since real-world AI systems often implement defenses regardless of whether an attack is present, we also evaluate their impact on clean images to examine potential performance degradation. Based on this setup, we define four robustness measures: robustness to intrinsic model errors, impact of adversarial countermeasures on hallucination performance, adversarial robustness without defense, and adversarial robustness with defense.

\subsubsection{Hallucination Benchmarks}

\paragraph{POPE.}
The POPE benchmark \cite{li2023evaluating} consists of a VQA dataset where models answer binary presence-related questions, such as ``\textit{Is there a \{OBJECT\} in the image?}" Performance is measured using Accuracy and F1 Score, while the ``Yes Ratio" metric assesses model bias. POPE includes three subsets: \textit{Random Subset}: This subset randomly chooses absent objects to form the questions. \textit{Popular Subset}: This subset selects frequently occurring objects. \textit{Adversarial Subset}: This subset is designed with co-occurring but absent objects to challenge models. It is important to clarify that adversarial subset in POPE does not imply any adversarial attacks. Following prior work, we sample 300 questions (50 images, 6 questions each, equally distributed between ``Yes" and ``No") per subset.

\paragraph{MME - Existence Subset.}
The MME benchmark's \textit{Existence} subset \cite{liang2024survey} evaluates an LVLM’s ability to determine object presence using questions similar to the POPE benchmark with binary responses. Performance is measured using Accuracy, Accuracy Plus (Acc+), which accounts for partially correct answers, and Total Score, which is the sum of Acc and Acc+.

\paragraph{AMBER - Generative Subset.}
For generative hallucination evaluation, we use the AMBER benchmark’s \textit{Generative} subset \cite{wang2023amber}, where models generate captions in response to ``\textit{Describe the image.}" Performance is evaluated with four specific metrics: \textit{CHAIR}, which measures the degree of hallucination in the generated captions by assessing the presence of objects not found in the annotated descriptions. \textit{Cover}, which evaluates the coverage of ground-truth objects in the generated captions, indicating how comprehensively the model describes the image content. \textit{Hal}, which quantifies the hallucination rate by calculating the ratio of objects mentioned in the generated captions that do not exist in the image. \textit{Cog}, which assesses whether models' hallucinations are similar to those in human cognition, utilizing a predefined set of hallucinatory objects. Given the computational demands of subsequent experiments involving adversarial attacks, we limit our evaluation to a sample of 50 images from this subset.

\subsubsection{Implementation Details}

\paragraph{Models in Hydra.}
To demonstrate Hydra’s adaptability across LVLMs, we evaluate it with multiple models. The plug-in LVLMs (specified with model sizes) include mPLUG-Owl (7B) \cite{ye2023mplug}, LLaVA-1.5 (7B) \cite{liu2024improved}, MiniGPT-4 (13B) \cite{zhu2023minigpt}, and Qwen-VL-Chat (13B) \cite{Qwen-VL}. For the visual-language suite, we use a closed-set object detector, VLM, and VLP model: DETR (42M) \cite{carion2020end}, Paligemma (3B) \cite{beyer2024paligemma}, and $\text{BLIP}_{\text{vqa}}$ (385M) \cite{li2022blip}, respectively. Hydra’s agent backbone is Llama-3.2-Vision-Instruct (11B) \cite{llama32vision}, which is also used as a captioning model for our visual-language suite. Importantly, the agent does not have direct access to images during reasoning and relies solely on textual inputs. 

\paragraph{Dehallucination Baselines.}
For post-hoc dehallucination baselines, we compare Hydra with two SOTA methods: Woodpecker \cite{yin2024woodpecker} and LogicCheckGPT \cite{wu2024logical}. To ensure a fair comparison, we implement both using Llama-3.2-Vision-Instruct (11B) as the correction model instead of GPT-3.5-turbo in the original works.

\paragraph{Adversarial Attacks.}
We apply two advanced attacks targeting the LVLM joint embedding space on the images of hallucination benchmarks. The first attack, Adversarial Illusions \cite{bagdasaryan2024adversarial}, is a white-box attack using a VLP model, ImageBind \cite{girdhar2023imagebind}, as the surrogate model with an attack budget of $\epsilon = \frac{16}{255}$. It generates targeted perturbations that align nonexistent objects with image representations, leveraging hallucinatory object annotations from both discriminative and generative benchmarks. The second attack, AttackVLM \cite{zhao2023evaluating}, uses a transfer- and query-based strategy (MF-ii + MF-tt) with an attack budget of $\epsilon = \frac{8}{255}$. We employ a VLP model, Unidiffuser \cite{bao2023one}, as the surrogate model and construct adversarial examples by pairing COCO captions \cite{chen2015microsoft} with images generated by Stable Diffusion \cite{rombach2022high} that contain mismatched objects. Both attacks have demonstrated strong transferability across various VLMs in their studies.

\paragraph{Adversarial Defense.}
For adversarial defense, we apply two classic image preprocessing methods on the images of hallucination benchmarks: JPEG compression \cite{dziugaite2016study} and feature squeezing \cite{xu2017feature}. These methods are widely used in prior studies due to their simplicity, ease of implementation, and effectiveness. JPEG compression (quality=50) reduces high-frequency noise while preserving essential image features, striking a balance between fidelity and robustness. Feature squeezing (bit-depth=4) mitigates adversarial effects by limiting pixel precision, forcing models to rely on high-level structures rather than fine-grained perturbations. These defenses provide a practical trade-off between computational efficiency and adversarial robustness.



\subsection{Results Analysis}


\subsubsection{Robustness to Intrinsic Model Errors}

Our results in Table \ref{tab:result1} demonstrate that Hydra outperforms all SOTA dehallucination methods across POPE subsets and multiple LVLMs, achieving the highest accuracy (average 90+\%) while maintaining a balanced Yes Ratio (Yes\%) around 50\%, i.e., the model's bias is more balanced. When applied to mPLUG-Owl, Hydra improves accuracy by over 40\%, also significantly surpassing both Woodpecker and LogicCheckGPT on the POPE-Random subset. On LLaVA-1.5, Hydra stabilizes accuracy, preventing the substantial drops (POPE-Random to POPE-Popular subset) observed without dehallucination methods applied. Similarly, on MiniGPT-4, Hydra achieves 20\% higher accuracy than the baseline. For Qwen-VL-Chat, Hydra outperforms Woodpecker by 20\% on the POPE-Popular subset, maintaining consistent accuracy across subsets, unlike Woodpecker, which fluctuates. Beyond accuracy, Hydra reduces the Yes Ratio, mitigating overconfidence. Models without dehallucination methods exhibit inflated values (e.g., mPLUG-Owl reaching 99\%), whereas Hydra significantly lowers them while preserving performance. These results confirm Hydra’s superior hallucination mitigation and stability across different LVLMs and POPE subsets.

\begin{table}[t]
    \footnotesize
    \centering
    \renewcommand{\arraystretch}{1}
    \begin{tabular}{llccc|ccc|ccc}
        \toprule
        \multicolumn{2}{c}{} & \multicolumn{3}{c}{\textbf{POPE-Ran}} & \multicolumn{3}{c}{\textbf{POPE-Pop}} & \multicolumn{3}{c}{\textbf{POPE-Adv}} \\
        \cmidrule(lr){3-5} \cmidrule(lr){6-8} \cmidrule(lr){9-11}
        Model & Dehallucination & Acc & F1 & Yes\% & Acc & F1 & Yes\% & Acc & F1 & Yes\% \\
        \midrule
        \multirow{4}{*}{mPlug-Owl} 
        & None & 54 & 68.5 & 96 & 51.3 & 67.3 & 98.7 & 51 & 67.1 & 99 \\
        & Woodpecker & 87.3 & 86.9 & 46.7 & 89 & 88.5 & 45.7 & 85 & 84.9 & 49 \\
        & LogicCheckGPT & 89.7 & 88.9 & 43 & 82.7 & 82.3 & 48 & 80.3 & 80.2 & 49.7 \\
        & Hydra & \textbf{94.7} & \textbf{94.4} & 44.7 & \textbf{92} & \textbf{91.8} & 48 & \textbf{90.3} & \textbf{90.2} & 49 \\
        \midrule
        \multirow{4}{*}{LLaVA-1.5} 
        & None & 90.7 & 91.9 & 54.7 & 83 & 84.6 & 60.3 & 82.3 & 84.4 & 63 \\
        & Woodpecker & 82.7 & 83.1 & 52.7 & 85 & 85.4 & 53 & 81.3 & 82.2 & 54.7 \\
        & LogicCheckGPT & 87 & 85.2 & 37.7 & 85.3 & 83.5 & 38.7 & 84 & 82.2 & 40 \\
        & Hydra & \textbf{94.3} & \textbf{94.0} & 45 & \textbf{93.7} & \textbf{93.4} & 45.7 & \textbf{91.7} & \textbf{91.5} & 47.7 \\
        \midrule
        \multirow{4}{*}{MiniGPT-4} 
        & None & 75 & 76.2 & 55 & 72 & 75.2 & 62.7 & 70 & 74.3 & 66.7 \\
        & Woodpecker & 82.3 & 81.8 & 47 & 80.3 & 80.7 & 51.7 & 76.3 & 76.6 & 51 \\
        & LogicCheckGPT & 74.7 & 68.6 & 30.7 & 77.3 & 71.4 & 29.3 & 72.7 & 67.7 & 34.7 \\
        & Hydra & \textbf{93.7} & \textbf{93.4} & 45.7 & \textbf{93.3} & \textbf{93.0} & 45.3 & \textbf{92} & \textbf{91.8} & 48 \\
        \midrule
        \multirow{4}{*}{Qwen-VL-Chat} 
        & None & 90.3 & 89.9 & 45.7 & 86 & 86 & 50 & 85.7 & 85.7 & 50.3 \\
        & Woodpecker & 81 & 82.1 & 56.3 & 72 & 75.9 & 66 & 76.7 & 79.3 & 62.7 \\
        & LogicCheckGPT & 80.3 & 75.5 & 30.3 & 82.3 & 78.5 & 32.3 & 82 & 78.5 & 28 \\
        & Hydra & \textbf{94} & \textbf{93.6} & 44 & \textbf{92.7} & \textbf{92.3} & 45.3 & \textbf{92.3} & \textbf{91.9} & 45 \\
        \bottomrule
        \\
    \end{tabular}
    \caption{Performance of LVLMs with dehallucination methods across POPE subsets. Abbreviations: POPE-Random Subset (Ran), POPE-Popular Subset (Pop), and POPE-Adversarial Subset (Adv). The best results in each model are \textbf{bolded}.}
    \label{tab:result1}
\end{table}

\subsubsection{Impact of Adversarial Defenses on Dehallucination}

Table \ref{tab:result2} shows that Hydra enhances adversarial robustness while maintaining high accuracy across POPE subsets, notably achieving 96.2\% accuracy with JPEG compression on MiniGPT-4. In AMBER, Hydra significantly reduces hallucinations (\textit{Hal}) and cognitive errors (\textit{Cog}) but at the cost of lower object coverage (\textit{Cover}), highlighting a trade-off between robustness and ground-truth retention. While image preprocessing defenses mitigate adversarial errors, they introduce degradation that affects dehallucination consistency. Without defenses, Hydra remains stable across POPE and AMBER, reinforcing its resilience in clean settings. These results underscore the need to balance adversarial robustness with image integrity to optimize dehallucination performance.

\begin{table}[t]
    \footnotesize
    \centering
    \renewcommand{\arraystretch}{1.1}
    \begin{tabular}{@{}llc|ccc|cccc@{}}
        \toprule
        \multicolumn{2}{c}{} & & \multicolumn{3}{c}{\textbf{POPE} (Acc$\uparrow$)} & \multicolumn{4}{c}{\textbf{AMBER-Generative}} \\
        \cmidrule(lr){4-6} \cmidrule(lr){7-10}
        Model & Defense & Hydra & \textbf{Ran} & \textbf{Pop} & \textbf{Adv} & CH$\downarrow$ & Cov$\uparrow$ & Hal$\downarrow$ & Cog$\downarrow$ \\
        \midrule
        \multirow{3}{*}{MiniGPT-4} 
        & None & \texttimes & 75.0 & 72.0 & 70.0 & 16.3 & \textbf{66.2} & 84.0 & 12.6 \\
        &  & \checkmark & \textbf{93.7} & \textbf{93.3} & \textbf{92.0} & \textbf{8.7} & 52.7 & \textbf{26.0} & \textbf{1.6} \\
        \cmidrule(lr){2-10}
        & JPEG & \texttimes & 72.3 & 65.0 & 63.7 & 16.4 & \textbf{63.3} & 72.0 & 9.3 \\
        &  & \checkmark & \textbf{96.2} & \textbf{96.2} & \textbf{88.3} & \textbf{10.7} & 48.4 & \textbf{32.0} & \textbf{5.3} \\
        \cmidrule(lr){2-10}
        & FeatSq & \texttimes & 77.0 & 72.3 & 68.0 & 13.8 & \textbf{70.5} & 74.0 & 10.6 \\
        &  & \checkmark & \textbf{91.0} & \textbf{91.0} & \textbf{89.7} & \textbf{10.2} & 52.7 & \textbf{28.0} & \textbf{2.8} \\
        \midrule
        \multirow{3}{*}{Qwen-VL-Chat} 
        & None & \texttimes & 90.3 & 86.0 & 85.7 & 7.5 & \textbf{57.8} & 40.0 & 4.1 \\
        &  & \checkmark & \textbf{94.0} & \textbf{92.7} & \textbf{92.3} & \textbf{3.2} & 46.9 & \textbf{8.0} & \textbf{0.4} \\
        \cmidrule(lr){2-10}
        & JPEG & \texttimes & 88.0 & 88.3 & 87.0 & 5.5 & \textbf{56.0} & 28.0 & 1.2 \\
        &  & \checkmark & \textbf{91.0} & \textbf{90.0} & \textbf{90.0} & \textbf{6.9} & 46.9 & \textbf{20.0} & \textbf{0.0} \\
        \cmidrule(lr){2-10}
        & FeatSq & \texttimes & 90.3 & 89.7 & 89.3 & 9.5 & \textbf{56.7} & 30.0 & 2.0 \\
        &  & \checkmark & \textbf{92.7} & \textbf{92.0} & \textbf{90.3} & \textbf{4.1} & 50.2 & \textbf{12.0} & \textbf{1.2} \\
        \bottomrule
        \\
    \end{tabular}
    \caption{Results of applying adversarial defenses with and without Hydra across POPE and AMBER benchmarks in clean setting. Abbreviations: JPEG Compression (JPEG), Feature Squeezing (FeatSq); POPE-Random Subset (Ran), POPE-Popular Subset (Pop), and POPE-Adversarial Subset (Adv); AMBER's metric, CHAIR (CH); Cover (Cov). Lower ($\downarrow$) or higher ($\uparrow$) values indicate better performance. The best results in each defense are \textbf{bold}.}
    \label{tab:result2}
\end{table}

\subsubsection{Adversarial Robustness Without Defense}

\begin{figure}[t]
    \centering
    \begin{subfigure}{0.49\linewidth}
        \centering
        \includegraphics[width=\linewidth]{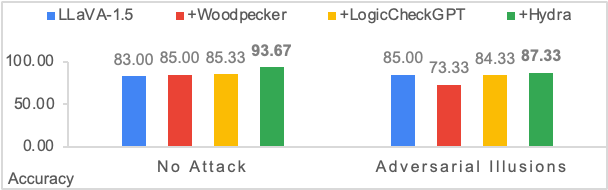}
        \label{fig:result3_1}
    \end{subfigure}
    \hfill
    \begin{subfigure}{0.49\linewidth}
        \centering
        \includegraphics[width=\linewidth]{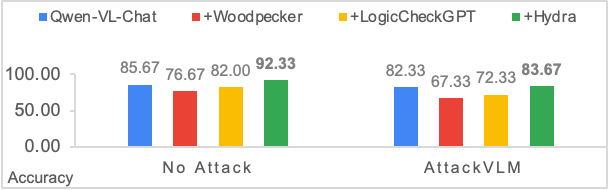}
        \label{fig:result3_2}
    \end{subfigure}
    \caption{The comparison of different dehallucination methods before and after applying adversarial attacks on POPE.}
    \label{fig:results3}
\end{figure}

Hydra exhibits strong adversarial robustness, outperforming plug-in LVLMs and SOTA dehallucination methods even without explicit defenses, as shown in Figure \ref{fig:results3}. On LLaVA-1.5 (evaluated on POPE-Popular subset), Hydra achieves 93.67\% accuracy before attack and maintains the highest performance (87.33\%) when it is attacked by Adversarial Illusions, showing better resilience than LogicCheckGPT (84.33\%) and Woodpecker (73.33\%). For Qwen-VL-Chat (evaluated on the POPE-Adversarial subset), Hydra leads with 92.33\% before the attack and remains the strongest performer (83.67\%) after being attacked by AttackVLM, outperforming LogicCheckGPT (72.33\%) and Woodpecker (67.33\%). These results confirm Hydra’s superior resistance to adversarial attacks while maintaining robustness against intrinsic errors.


\subsubsection{Adversarial Robustness With Defense}

Across the three tables (Table \ref{tab:result4}, \ref{tab:result5}, \ref{tab:result6}, Hydra demonstrates superior performance in both clean and adversarial settings, consistently outperforming state-of-the-art (SOTA) dehallucination methods such as LogicCheckGPT and Woodpecker. In POPE-Random subset (Table \ref{tab:result4}), Hydra achieves high accuracy and F1 scores across all defense configurations, indicating its robustness against Adversarial Illusions and AttackVLM attacks. Even under JPEG compression and Feature Squeezing, Hydra maintains strong dehallucination performance, significantly reducing the percentage of hallucinated responses compared to other methods. This suggests that Hydra effectively mitigates adversarial manipulations while preserving accuracy, making it a reliable choice across diverse defense strategies.

In MME-Existence (Table \ref{tab:result5}), Hydra exhibits remarkable resilience across different attack scenarios. When applied to mPlug-Owl and MiniGPT-4, it maintains high accuracy even under adversarial attacks, surpassing baseline models by a significant margin. Notably, Hydra sustains its robustness under Adversarial Illusions and AttackVLM conditions, demonstrating minimal performance drop (-5\% in Total) compared to clean settings, highlighting Hydra’s ability to generalize well across varying attack intensities. Interestingly, in both MME-Existence and AMBER (Table \ref{tab:result6}), we observe cases where performance improves after attack. This suggests that certain adversarial perturbations may unintentionally align with the VLM's reasoning patterns, allowing it to bypass attacks. The underlying cause of this phenomenon remains an open question, warranting further investigation.

Table \ref{tab:result6} further confirms Hydra’s effectiveness on AMBER-Generative, where it significantly reduces baseline VLMs' hallucination rate (\textit{Hal}) and cognitive errors (\textit{Cog}) across all defenses. Even with Adversarial Illusions and AttackVLM perturbations, Hydra improves CHAIR scores and reduces hallucinations comparing with baseline VLMs. However, its \textit{Cover} metric does not outperform baseline VLMs, which is a natural trade-off. Since Hydra prioritizes hallucination reduction, it adopts a more conservative approach, making it less likely to falsely assert object existence. This cautious stance enhances robustness against hallucinations but inherently lowers coverage scores. Despite this, Hydra’s overall stability across models and benchmarks underscores its effectiveness as a robust and adaptive dehallucination framework.

\begin{table}[t]
    \footnotesize
    \centering
    \renewcommand{\arraystretch}{1.1}
    \begin{tabular}{ll|ccc|ccc|ccc}
        \toprule
        \multicolumn{11}{c}{\textbf{mPlug-Owl on POPE-Random}} \\
        \midrule
        \multicolumn{2}{c|}{} & \multicolumn{3}{c}{\textbf{No Attack}} & \multicolumn{3}{c}{\textbf{Adv. Illusions}} & \multicolumn{3}{c}{\textbf{AttackVLM}} \\
        \cmidrule(lr){3-5} \cmidrule(lr){6-8} \cmidrule(lr){9-11}
        Defense & Dehallucination & Acc & F1 & Yes\% & Acc & F1 & Yes\% & Acc & F1 & Yes\% \\
        \midrule
        \multirow{4}{*}{None} 
            & None & 54.0 & 68.5 & 96.0 & 51.3 & 67.3 & 98.7 & 51.0 & 67.1 & 99.0 \\
        & Woodpecker & 87.3 & 86.9 & 46.7 & 77.7 & 74.3 & 37.0 & 84.0 & 82.1 & 39.3 \\
        & LogicCheckGPT & 89.7 & 88.9 & 43.0 & \textbf{86.0} & \textbf{84.9} & 43.7 & \textbf{88.7} & \textbf{87.9} & 43.3 \\
        & Hydra & \textbf{94.7} & \textbf{94.4} & 44.7 & \textbf{86.0} & 84.4 & 40.0 & \textbf{88.7} & 87.5 & 40.7 \\
        \midrule
        \multirow{4}{*}{JPEG} 
            & None & 51.3 & 67.3 & 98.7 & 51.3 & 67.3 & 98.7 & 51.0 & 67.1 & 99.0 \\
        & Woodpecker & 77.3 & 72.8 & 33.3 & 77.7 & 73.4 & 35.3 & 81.7 & 78.9 & 37.0 \\
        & LogicCheckGPT & 82.3 & 81.0 & 43.0 & 84.7 & 83.7 & 44.0 & 86.7 & 85.9 & 44.7 \\
        & Hydra & \textbf{92.0} & \textbf{91.5} & 44.0 & \textbf{86.3} & \textbf{84.5} & 38.3 & \textbf{87.7} & \textbf{86.3} & 39.7 \\
        \midrule
        \multirow{4}{*}{FeatSq} 
            & None & 51.3 & 67.3 & 98.7 & 53.0 & 68.0 & 97.0 & 51.3 & 67.3 & 98.7 \\
        & Woodpecker & 87.3 & 87.2 & 48.7 & 78.7 & 75.8 & 38.0 & 82.0 & 79.6 & 38.0 \\
        & LogicCheckGPT & 84.7 & 83.2 & 41.3 & 82.3 & 80.6 & 41.0 & 84.3 & 82.4 & 39.0 \\
        & Hydra & \textbf{92.3} & \textbf{91.9} & 45.0 & \textbf{85.0} & \textbf{83.0} & 38.3 & \textbf{89.0} & \textbf{88.0} & 41.7 \\
        \bottomrule
        \\
    \end{tabular}
    \caption{The performance comparison of different adversarial defenses and dehallucination methods across various attacks on POPE-Random subset. Abbreviations: JPEG Compression (JPEG), Feature Squeezing (FeatSq). The best result, comparing different dehallucination methods within the same defense, is highlighted in \textbf{bold}.}
    \label{tab:result4}
\end{table}

\begin{table}[t]
    \footnotesize
    \centering
    \renewcommand{\arraystretch}{1.1}
    \begin{tabular}{llc|ccc|ccc|ccc}
        \toprule
        \multicolumn{12}{c}{\textbf{MME-Existence}} \\
        \midrule
        \multicolumn{3}{c}{} & \multicolumn{3}{c}{\textbf{No Attack}} & \multicolumn{3}{c}{\textbf{Adv. Illusions}} & \multicolumn{3}{c}{\textbf{AttackVLM}} \\
        \cmidrule(lr){4-6} \cmidrule(lr){7-9} \cmidrule(lr){10-12}
        Model & Defense & Hydra & Acc & Acc+ & Total & Acc & Acc+ & Total & Acc & Acc+ & Total \\
        \midrule
        \multirow{3}{*}{mPlug-Owl} 
        & None & \texttimes & 63.3 & 26.7 & 90 & 50.0 & 0.0 & 50 & 51.7 & 3.3 & 55 \\
        &   & \checkmark & \textbf{93.3} & \textbf{86.7} & \textbf{180} & \textbf{91.7} & \textbf{83} & \textbf{175} & \textbf{91.7} & \textbf{83.3} & \textbf{175} \\
        \cmidrule(lr){2-12}
        & JPEG & \texttimes & 50.0 & 0.0 & 50 & 50.0 & 0.0 & 50 & 50.0 & 0.0 & 50 \\
        &   & \checkmark & \textbf{91.7} & \textbf{83.3} & \textbf{175} & \textbf{91.7} & \textbf{83} & \textbf{175} & \textbf{88.3} & \textbf{76.7} & \textbf{165} \\
        \cmidrule(lr){2-12}
        & FeatSq & \texttimes & 58.3 & 16.7 & 75 & 51.7 & 3.3 & 55 & 51.7 & 3.3 & 55 \\
        &   & \checkmark & \textbf{93.3} & \textbf{86.7} & \textbf{180} & \textbf{93.3} & \textbf{86.7} & \textbf{180} & \textbf{93.3} & \textbf{86.7} & \textbf{180} \\
        \midrule
        \multirow{3}{*}{MiniGPT-4} 
        & None & \texttimes & 73.3 & 53.3 & 126.7 & 71.7 & 53.3 & 125 & 75.0 & 60.0 & 135 \\
        &   & \checkmark & \textbf{91.7} & \textbf{83.3} & \textbf{175} & \textbf{90} & \textbf{80.0} & \textbf{170} & \textbf{88.3} & \textbf{76.7} & \textbf{165} \\
        \cmidrule(lr){2-12}
        & JPEG & \texttimes & 70.0 & 56.7 & 126.7 & 78.3 & 63.3 & 141.7 & 71.7 & 53.3 & 125 \\
        &   & \checkmark & \textbf{91.7} & \textbf{83.3} & \textbf{175} & \textbf{90} & \textbf{80} & \textbf{170} & \textbf{88.3} & \textbf{76.7} & \textbf{165} \\
        \cmidrule(lr){2-12}
        & FeatSq & \texttimes & 75.0 & 60.0 & 135 & 65.0 & 43.3 & 108.3 & 70.0 & 50.0 & 120 \\
        &   & \checkmark & \textbf{95} & \textbf{90} & \textbf{185} & \textbf{86.7} & \textbf{73.3} & \textbf{160} & \textbf{90} & \textbf{80} & \textbf{170} \\
        \bottomrule
        \\
    \end{tabular}
    \scriptsize \caption{Adversarial robustness comparison on MME-Existence Subset. Evaluates adversarial defenses and Hydra across different attack settings for mPlug-Owl and MiniGPT-4. Abbreviations: JPEG Compression (JPEG), Feature Squeezing (FeatSq). The best result, comparing different dehallucination methods within the same defense, is highlighted in \textbf{bold}.}
    \label{tab:result5}
\end{table}

\begin{table*}[t]
    \scriptsize
    \centering
    \setlength{\tabcolsep}{4pt} 
    \renewcommand{\arraystretch}{1.2}
    \begin{tabular}{llc|cccc|cccc|cccc}
        \toprule
        \multicolumn{15}{c}{\textbf{AMBER-Generative}} \\
        \midrule
        \multicolumn{3}{c}{} & \multicolumn{4}{c}{\textbf{No Attack}} & \multicolumn{4}{c}{\textbf{Adversarial Illusions}} & \multicolumn{4}{c}{\textbf{AttackVLM}} \\
        \cmidrule(lr){4-7} \cmidrule(lr){8-11} \cmidrule(lr){12-15}
        Model & Defense & Hydra & CHAIR$\downarrow$ & Cover$\uparrow$ & Hal$\downarrow$ & Cog$\downarrow$ & CHAIR$\downarrow$ & Cover$\uparrow$ & Hal$\downarrow$ & Cog$\downarrow$ & CHAIR$\downarrow$ & Cover$\uparrow$ & Hal$\downarrow$ & Cog$\downarrow$ \\
        \midrule
        \multirow{9}{*}{mPlug-Owl} 
        & \multirow{2}{*}{None} 
            & \texttimes & 23.20 & \textbf{49.80} & 82.00 & 16.70 & 24.50 & \textbf{46.20} & 70.00 & 10.60 & 24.40 & \textbf{50.20} & 84.00 & 11.80 \\
        &   & \checkmark & \textbf{19.70} & 41.10 & \textbf{56.00} & \textbf{7.70} & \textbf{18.80} & 41.10 & \textbf{40.00} & \textbf{4.50} & \textbf{20.70} & 42.50 & \textbf{54.00} & \textbf{4.90} \\
        \cmidrule(lr){2-15}
        & \multirow{2}{*}{JPEG} 
            & \texttimes & 19.40 & \textbf{49.10} & 72.00 & 11.40 & \textbf{18.50} & \textbf{46.20} & 54.00 & 9.30 & \textbf{19.60} & \textbf{51.60} & 72.00 & 11.00 \\
        &   & \checkmark & \textbf{17.80} & 40.70 & \textbf{46.00} & \textbf{4.10} & 20.70 & 36.00 & \textbf{40.00} & \textbf{4.90} & 19.80 & 41.80 & \textbf{46.00} & \textbf{7.30} \\
        \cmidrule(lr){2-15}
        & \multirow{2}{*}{FeatSq} 
            & \texttimes & 21.50 & \textbf{50.20} & 92.00 & 13.40 & 19.50 & 44.70 & 70.00 & 10.20 & 19.90 & \textbf{47.30} & 70.00 & 12.20 \\
        &   & \checkmark & \textbf{20.10} & 41.80 & \textbf{42.00} & \textbf{5.30} & \textbf{13.40} & 38.90 & \textbf{32.00} & \textbf{3.30} & \textbf{18.40} & 40.70 & \textbf{42.00} & \textbf{6.10} \\
        \midrule
        \multirow{9}{*}{MiniGPT-4} 
        & \multirow{2}{*}{None} 
            & \texttimes & 16.30 & \textbf{66.20} & 84.00 & 12.60 & 17.70 & \textbf{64.00} & 74.00 & 11.40 & 16.40 & \textbf{61.50} & 78.00 & 13.40 \\
        &   & \checkmark & \textbf{8.7} & 52.70 & \textbf{26.00} & \textbf{1.60} & \textbf{14.50} & 45.80 & \textbf{38.00} & \textbf{1.60} & \textbf{13.50} & 45.50 & \textbf{40.00} & \textbf{3.70} \\
        \cmidrule(lr){2-15}
        & \multirow{2}{*}{JPEG} 
            & \texttimes & 16.40 & \textbf{63.30} & 72.00 & 9.30 & 21.00 & \textbf{58.50} & 72.00 & 14.20 & 18.40 & \textbf{63.30} & 66.00 & 13.00 \\
        &   & \checkmark & \textbf{10.70} & 48.40 & \textbf{32.00} & \textbf{5.30} & \textbf{9.90} & 44.70 & \textbf{24.00} & \textbf{4.10} & \textbf{13.80} & 45.80 & \textbf{32.00} & \textbf{2.80} \\
        \cmidrule(lr){2-15}
        & \multirow{2}{*}{FeatSq} 
            & \texttimes & 13.80 & \textbf{70.50} & 74.00 & 10.60 & 15.60 & \textbf{68.70} & 60.00 & 9.80 & 19.60 & \textbf{64.40} & 68.00 & 14.60 \\
        &   & \checkmark & \textbf{10.20} & 52.70 & \textbf{28.00} & \textbf{2.80} & \textbf{8.40} & 50.20 & \textbf{22.00} & \textbf{1.20} & \textbf{11.50} & 46.90 & \textbf{28.00} & \textbf{2.40} \\
        \midrule
        \multirow{9}{*}{Qwen-VL-Chat} 
        & \multirow{2}{*}{None} 
            & \texttimes & 7.50 & \textbf{57.80} & 40.00 & 4.10 & 10.40 & \textbf{51.60} & 34.00 & 3.30 & 7.00 & \textbf{50.20} & 30.00 & 3.30 \\
        &   & \checkmark & \textbf{3.20} & 46.90 & \textbf{8.00} & \textbf{0.40} & \textbf{3.00} & 41.80 & \textbf{24.00} & \textbf{0.40} & \textbf{4.50} & 44.40 & \textbf{10.00} & \textbf{0.00} \\
        \cmidrule(lr){2-15}
        & \multirow{2}{*}{JPEG} 
            & \texttimes & 5.50 & \textbf{56.00} & 28.00 & 1.20 & 10.90 & \textbf{40.70} & 34.00 & 1.60 & \textbf{7.60} & \textbf{41.80} & \textbf{18.00} & 1.20 \\
        &   & \checkmark & \textbf{6.90} & 46.90 & \textbf{20.00} & \textbf{0.00} & \textbf{5.80} & 39.60 & \textbf{16.00} & 0.40 & 8.20 & 39.30 & 22.00 & \textbf{0.80} \\
        \cmidrule(lr){2-15}
        & \multirow{2}{*}{FeatSq} 
            & \texttimes & 9.50 & \textbf{56.70} & 30.00 & 2.00 & 12.80 & \textbf{49.10} & 42.00 & 3.30 & 5.70 & \textbf{52.00} & 26.00 & 1.60 \\
        &   & \checkmark & \textbf{4.10} & 50.20 & \textbf{12.00} & \textbf{1.20} & \textbf{10.10} & 44.00 & \textbf{26.00} & \textbf{1.20} & \textbf{2.50} & 44.70 & \textbf{8.00} & \textbf{0.00} \\
        \bottomrule
    \end{tabular}
    \scriptsize \caption{Adversarial robustness comparison on AMBER - Generative subset. It compares different adversarial defenses with and without Hydra across multiple attacks and models.Abbreviations: JPEG Compression (JPEG), Feature Squeezing (FeatSq). Lower ($\downarrow$) or higher ($\uparrow$) values indicate better performance. The best result, comparing with and without applying Hydra under the same defense and metric, is highlighted with in \textbf{bold}.}
    \label{tab:result6}
\end{table*}

\section{Conclusion}


In this work, we introduced Hydra, an adaptive agentic reasoning framework designed to enhance the robustness of VLMs against both adversarial perturbations and intrinsic model errors. By integrating structured multi-step reasoning with an action-critique loop, Hydra effectively refines VLM outputs and mitigates object-level hallucinations. Our empirical evaluation across multiple SOTA VLMs, hallucination benchmarks, and adversarial attack scenarios demonstrates that Hydra surpasses existing dehallucination methods while maintaining resilience against adversarial manipulations. Notably, our results indicate that Hydra’s modular design allows for improved factual alignment and interpretability, offering a training-free and model-agnostic approach to VLM robustness enhancement.

Beyond improving robustness to intrinsic errors, Hydra’s agentic decision-making mechanism exhibits strong adversarial resistance, even without explicit adversarial defenses. The framework demonstrates that leveraging a structured visual-language suite with small-scale models (sizes under 3B) can yield significant improvements, particularly under computational constraints. Moreover, our findings highlight the potential of agent-based frameworks to unify robustness against both adversarial perturbations and generative errors, paving the way for future research in adaptive defense mechanisms for multimodal AI systems.

\paragraph{Limitations.}
Nevertheless, Hydra is not without limitations. The computational complexity of iterative agentic reasoning largely increases inference time, which may hinder real-time applications. Additionally, while Hydra effectively mitigates object-level hallucinations, extending this framework to attribute- and relationship-level hallucinations remains an open challenge. Furthermore, current adversarial defenses in VLMs primarily focus on pre-processing techniques or adversarial training, while there remains a critical gap in defense strategies targeting the joint embedding space of VLMs—an avenue that requires further exploration.

\paragraph{Future.}
We suggest investigate more efficient multimodal agentic architectures that retain robustness while reducing computational overhead, such as exploring the recent advancement of scaling test-time compute. Additionally, exploring adaptive defense mechanisms tailored to multimodal embedding spaces could provide new pathways for securing VLMs against both adversarial and intrinsic errors. By bridging the gap between agentic AI and adversarial robustness, Hydra serves as a step toward trustworthy, interpretable, and resilient vision-language reasoning systems.

\section*{Acknowledgments}
This work was supported in part by the U.S. Military Academy (USMA) under Cooperative Agreement No. W911NF-23-2-0108. The views and conclusions expressed in this paper are those of the authors and do not reflect the official policy or position of the U.S. Military Academy, U.S. Army, U.S. Department of Defense, or U.S. Government.

\bibliographystyle{unsrt}  
\bibliography{references}  


\end{document}